\documentclass{article}

\usepackage[final,nonatbib]{neurips_2020}

\usepackage[utf8]{inputenc} 
\usepackage[T1]{fontenc}    
\usepackage{hyperref}       
\usepackage{url}            
\usepackage{booktabs}       
\usepackage{amsfonts}       
\usepackage{nicefrac}       
\usepackage{microtype}      

\usepackage{latexsym,amssymb,amstext,amsfonts,amsmath,graphicx,setspace,mathtools,bm}
\usepackage{subcaption}
\usepackage[colorinlistoftodos]{todonotes}
\usepackage{enumerate} 
\usepackage{paralist}
\usepackage{float}
\usepackage{array}

\title{Harmonization and the Worst Scanner Syndrome}

\author{%
  Daniel Moyer and Polina Golland\\
  Computer Science and Artificial Intelligence Lab\\
  Massachusetts Institute of Technology\\
  Cambridge (MA) 02139, USA \\
  \texttt{\{dmoyer,polina\}@csail.mit.edu}
}

\begin{document}

\maketitle

\begin{abstract}
We show that for a wide class of harmonization/domain-invariance schemes several undesirable properties are unavoidable. If a predictive machine is made invariant to a set of domains, the accuracy of the output predictions (as measured by mutual information) is limited by the domain with the \emph{least} amount of information to begin with. If a real label value is highly informative about the source domain, it cannot be accurately predicted by an invariant predictor. These results are simple and intuitive, but we believe that it is beneficial to state them for medical imaging harmonization.  
\end{abstract}



All images in medical imaging have a scanner-bias. We receive images $\mathcal{X}$ from e.g. an MRI machine, and we would like to predict a label $\mathcal{Y}$, which may be a disease state, or a tumor location, or a tissue label map. The images are in part dependent on these labels, but are also dependent on the equipment that collected them, the scanner/site variables $\mathcal{S}$. En masse, the biases in equipment (or inhomogeneity in the populations put into one piece of equipment versus another) can be a predictive of the labels $\mathcal{Y}$ in a static dataset.

We know from intuition that most labels are not mediated by the scanner, that using the MRI machine in Room A opposed to Room B does not \emph{actually} change the disease risk\footnote{We use ``disease risk'' to refer to actual risk of having a disease, opposed to the diagnosis label assigned by a physician. Diagnosis may be mediated by scanner, as it is assigned possibly in consultation with radiologists, whose ``input data'' is from the scanner. This varies by context, and care should be taken when assessing whether diagnostic data are influenced by scanner.} of a patient; a reasonable clinician would not diagnose a patient based on which machine collected the images. Recent methods using neural networks provide state-of-the-art predictive performance, but unfortunately are also flexible enough to learn the scanner information \cite{glocker2019machine}. Because scanner information is often predictive within the training data, a suitably flexible method will predict labels based off of that information, exactly the opposite of the ``reasonable clinician'' case.

It is thus useful to find predictors of $y$ from $x$ that are invariant to collection scanner/site $s$. These methods are collectively referred to as harmonized or domain-invariant predictors, and substantial effort has been put forth to construct them for medical imaging applications \cite{dewey2020disentangled,yang2019unsupervised,ilse2020diva,moyer2020scanner,aslani2020scanner,dinsdale2020unlearning,kamnitsas2017unsupervised,lafarge2019learning}.

In this manuscript we first characterize a large class of these methods (``Invariant Representation--Data Processing Inequality'' or IR-DPI methods), and then show that every method in this class has specific adverse properties, namely that if a method from this class is actually domain-invariant, then its predictive power is bounded by the least-informative domain. We also show that for subsets of the target domain $\mathcal{Y}$ which are highly correlated with specific values of $\mathcal{S}$, accuracy under invariance is necessarily lost. We then discuss implications for the application of these methods.


This paper is not meant to dissuade researchers from work on IR-DPI class methods, but instead highlight specific error modes, around which results and expected behaviors should be framed. 

\section{Invariant Representation--Data Processing Inequality methods}


Consider a Markov chain $(\mathcal{Y},\mathcal{S}) \rightarrow \mathcal{X} \rightarrow \mathcal{\hat{Y}}$, where we observe $x,s$ (and $y$ for a training dataset), and would like construct $\hat{y}$ that predicts $y$, minimizing a risk e.g. $R(\hat{y}) = \|y - \hat{y}\|$. We would further like to learn a harmonized/domain-invariant predictor $\hat{y}$ that fits the constraint
\begin{align}
p(\hat{y}|x,s) = p(\hat{y} |x,s') \text{ for all }s,s' \in \mathcal{S}. 
\end{align}
This is equivalent to the constraint that $\hat{y}$ be uninformative of $s$, i.e.
\begin{align}
I(\hat{y},s) = 0.
\end{align}
Many methods relax this hard constraint to a trade-off between minimizing risk and minimizing site information, resulting in objective functions like
\begin{align}
\mathcal{L}[\hat{y}] = R(\hat{y}) - \lambda I(\hat{y},s),
\end{align}
where $\lambda$ mediates the trade-off between risk and invariance.

A popular method to achieve this introduces the intermediate representation $\mathcal{Z}$, where prediction takes the path $\mathcal{X} \rightarrow \mathcal{Z} \rightarrow \hat{\mathcal{Y}}$. By the data processing inequality, $I(z,s) \geq I(\hat{y},s)$, so by removing information about $\mathcal{S}$ from the learned representation $\mathcal{Z}$ we can effectively ensure invariant predictions $\hat{y}$. Predictive invariance is then achieved by minimization of $R(x)$ at the same time as $I(z,s)$. We refer to this class as Invariant Representation--Data Processing Inequality (IR-DPI) methods.

Adversarially robust methods such as Domain Adversarial Neural Networks \cite{ganin2016domain}, and those based off DANNs \cite{aslani2020scanner,dinsdale2020unlearning,kamnitsas2017unsupervised,lafarge2019learning} are subsumed by this class of methods, as are alternative methods for bounding or penalizing $I(z,s)$ \cite{ilse2020diva,moyer2020scanner} including many ``disentanglement'' methods \cite{dewey2020disentangled,yang2019unsupervised}. See \cite{moyer2018invariant} for a demonstration of the equivalence between adversarial and bounding/penalizing methods.

Moreover, invariant predictors $\hat{y}$ that constrain their outputs to not have site information are also a special case of IR-DPI methods. This can be seen by letting $\mathcal{Z} = \mathcal{Y}$.

\subsection{Worst Scanner Syndrome}

\textbf{Proposition 1:} Given a IR-DPI class predictor $\hat{y}$ that achieves invariance (i.e. $I(z,s) = 0$), and where $s$ and $y$ are independent ($I(s,y) = 0$), the following holds:
\begin{align}
I(y,\hat{y}) \leq \min_{s' \in \mathcal{S}} I(y,x|s=s')
\end{align}

\textbf{Proof:}
By the data processing inequality we have
\begin{align}
I(y,\hat{y}) \leq I(y,z) 
\end{align}
Without loss of generality, choose an $s' \in \mathcal{S}$. Writing out the definition of $I(y,z|s=s')$, we have
\begin{align}
I(y,z|s=s') & = \mathbb{E}_{y,z|s=s'}[\log p(y,z|s=s') - \log p(y|s=s')p(z|s=s') ] \\
& = \mathbb{E}_{y,z|s=s'}[\log p(y|s=s')p(z|y,s=s') - \log p(y|s=s')p(z|s=s') ] \\
& = \mathbb{E}_{y,z|s=s'}[\log p(z|y,s=s') - \log p(z|s=s') ] \\
& = \mathbb{E}_{y,z}[\log p(z|y) - \log p(z) ] \label{eq:inv} \\
& = I(y,z)
\end{align}

Eq.~\ref{eq:inv} follows from invariance. To complete the proof, let $s' = \arg\min_{s^*} I(y,z|s=s^*)$.

Proposition 1 implies that if a method is truly $s$-invariant, i.e. $I(z,s) = 0$, then the information in a predictor $\hat{y}$ about its target $y$ is bounded by the amount of information in the \emph{least} informative site. The ``worse scanner'' mediates the power of every IR-DPI method. If data are included from a highly informative scanner (high resolution, high SNR, etc.) alongside other normal scanners, the IR-DPI method will only be able to use information shared between all scanners. Similarly, if a bad scanner is included in the training set (e.g. a scanner that outputs mostly noise), the method will perform no better than if all scanners had been of that bad scanner quality. We call this phenomenon ``worst scanner syndrome''.

The $I(s,y) = 0$ case is clearly restrictive, but does not imply that site does not mediate scan quality. Instead, it is the constraint that scan quality is not correlated with label without regard to $x$; these may still be entangled in $x$, where a lower quality scan may delete information about $y$.

\subsection{Cases with Correlation between $\mathcal{S}$ and $\mathcal{Y}$}

We conjecture that a similar property should still hold for the $I(s,y) > 0$ case. This becomes difficult to show constructively, but we can still reason about these cases through examples and contradiction.



\textbf{Proposition 2:} Suppose that certain values $y'$ of $\mathcal{Y}$ are only observed at a specific site $s'$. If $I(z,s) = 0$ (i.e. our predictor is perfectly invariant), then it cannot accurately predict $y'$, or it necessary predicts other $y'$ values incorrectly as $y'$.

\textbf{Demonstration: } Suppose that it did accurately output $\hat{y} = y'$. If other values are not incorrectly predicted as $y'$, values of $z$ for which $\hat{y} = y'$ would be indicative of $s = s'$. Thus, $I(z,s) > 0$.

This becomes more complicated if we allow $I(z,s)$ to be small but non-zero, or consider larger but not covering subsets of $\mathcal{S}\backslash s'$, but the same principle applies, following from the same logic; for values of $y$ that are distributionally dissimilar between different values of $s$, reducing $I(z,s)$ implies reducing accuracy for that range of $y$.


\subsection{Generative Harmonization}

A number of methods propose Image-to-Image mappings from one scanner context to another \cite{ilse2020diva,moyer2020scanner}. Though we have framed the above discussion of $\mathcal{Y}$ as a dependent value, the same results can be applied if $\mathcal{Y}$ coincides with the image domain $\mathcal{X}$.

It may be tempting to search for $\mathcal{Y}$ or $\mathcal{Z}$ which are ``site-less'' images, i.e. images that have no scanner information. While we cannot show that this is impossible, it is not entirely sound. There are no examples of images without scanner biases, as all images were taken using a scanner. This is asking the quotient space of images over scanners to itself be a set of images, which may be difficult.

\section{Implications and Applications}

It is not the purpose of this paper to dissuade either practitioners or methodologists from IR-DPI class methods; far from it, many of the current methods in the literature fall into this class, and it is clear that a component of generalization in medical imaging requires accounting for site-wise signal. However, care must be taken when training and assessing these methods.

IR-DPI methods can prevent overfitting to site-wise signals, a form of overfitting which has increasing likelihood with the use of flexible function approximators (e.g. neural networks). Removal of this error-mode is critical for generalization, but it is unreasonable to expect methods to generalize to noisy data and still be just as performant. As we have shown above, forcing outputs to be invariant to all scanners implies worst-case scanner accuracy. Points on the curve of solutions inbetween unconstrained and invariant are likely more optimal.

If a training dataset and test dataset are not subject to distributional shifts, it is unlikely that methods can both be invariant and perform better than their unconstrainted counterpart unless other significant overfitting (not related to site-wise signal) has taken place. IR-DPI methods have compressive regularizers, so information in $\mathcal{Z}$ about $\mathcal{X}$ is being erased. Optimally this is information about $\mathcal{S}$, but nevertheless less information implies less predictive power. Compressive regularization can aid in generalization in general \cite{tishby2015deep}, but it is unclear if this is always the case.



Other forms of robust training that avoid IR-DPI may also be useful. Data augmentation naively constructs equivalence classes in the training set through enumeration (e.g. equivariant labels under rotations, translations, contrasts \cite{billot2020learning}, etc.), and is outside of the IR-DPI class. Even though the site variable cannot easily be augmented, artificially modifying collection conditions may reduce the method's ability to learn the site variable from the image, which in turn would reduce its propensity to learn site biases.

The objective of domain generalization is more general than invariance \cite{dou2019domain}. While invariant representation/feature learning is certainly a useful component of domain generalization, performant methods need not exhibit true invariance to be robust to domain shifts. Moreover domain invariance may be an insufficient desiderata. A completely blind introduction of new sources is unrealistic, so a better adaptation to shifts may look for test-time understanding of the target domain.

\subsection*{Broader Impacts}

This work presents theory results, and does not present any direct foreseeable consequence to the general public.

\subsection*{Acknowledgements}

This work was funded by {NIH NIBIB NAC P41EB015902}, {NIH NICHD R01HD100009}, {MIT Lincoln Lab}, and {Takeda}.

\begin{ack}
\end{ack}


\bibliography{main.bib}
\bibliographystyle{abbrv}

\end{document}